\documentclass[10pt,twocolumn,letterpaper]{article}

\usepackage{cvpr}
\usepackage{times}
\usepackage{epsfig}
\usepackage{graphicx}
\usepackage{amsmath}
\usepackage{amssymb}
\usepackage{multirow}

\usepackage[pagebackref=true,breaklinks=true,letterpaper=true,colorlinks,bookmarks=false]{hyperref}

\cvprfinalcopy 

\def\cvprPaperID{****} 

\ifcvprfinal\fi
\begin{document}

\title{Temporal Action Localization in Untrimmed Videos via Multi-stage CNNs }

\author{Zheng Shou, Dongang Wang, and Shih-Fu Chang\\
Columbia University\\
New York, NY, USA\\
{\tt\small \{zs2262,dw2648,sc250\}@columbia.edu}
}

\clearpage\maketitle
\thispagestyle{empty}

\begin{abstract}
We address temporal action localization in untrimmed long videos. This is important because videos in real applications are usually unconstrained and contain multiple action instances plus video content of background scenes or other activities. To address this challenging issue, we exploit the effectiveness of deep networks in temporal action localization via three segment-based 3D ConvNets: (1) a \textbf{proposal} network identifies candidate segments in a long video that may contain actions; (2) a \textbf{classification} network learns one-vs-all action classification model to serve as initialization for the localization network; and (3) a \textbf{localization} network fine-tunes the learned classification network to localize each action instance. We propose a novel loss function for the localization network to explicitly consider temporal overlap and achieve high temporal localization accuracy. In the end, only the proposal network and the localization network are used during prediction. On two large-scale benchmarks, our approach achieves significantly superior performances compared with other state-of-the-art systems: mAP increases from 1.7$\%$ to 7.4$\%$ on MEXaction2 and increases from 15.0$\%$ to 19.0$\%$ on THUMOS 2014.


\end{abstract}

\section{Introduction}

\begin{figure*}[t]
\centering
\includegraphics[width=\textwidth]{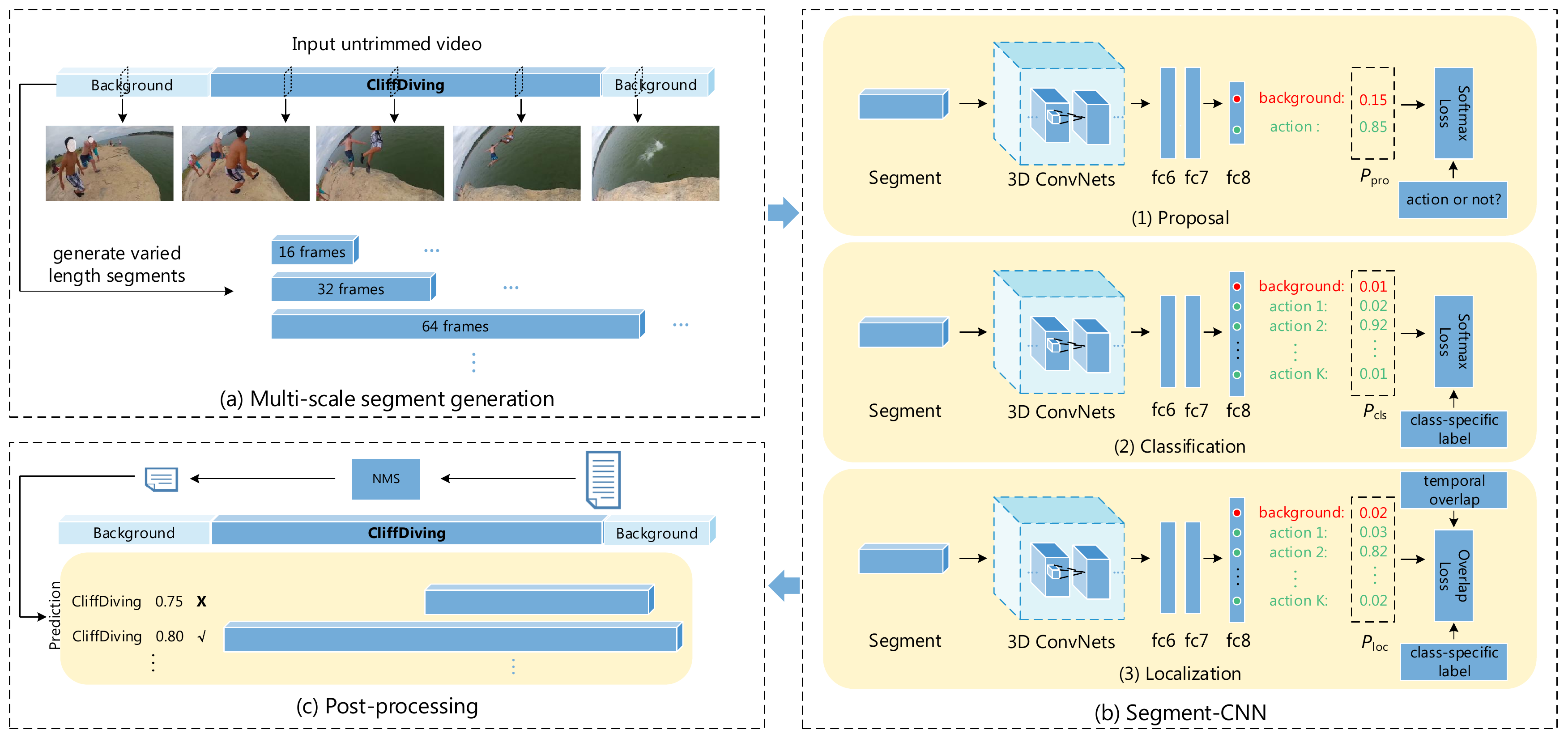}
\caption{Overview of our framework. (a) Multi-scale segment generation: given an untrimmed video, we generate segments of varied lengths via sliding window; (b) Segment-CNN: the \textbf{proposal} network identifies candidate segments, the \textbf{classification} network trains an action recognition model to serve as initialization for the localization network, and the \textbf{localization} network localizes action instances in time and outputs confidence scores; (c) Post-processing: using the prediction scores from the localization network, we further remove redundancy by NMS to obtain the final results. During training, the classification network is first learned and then used as initialization for the localization network. During prediction, only the proposal and localization networks are used.}
\label{framework}
\end{figure*}

Impressive progress has been reported in recent literature for action recognition \cite{survey1,survey2,survey3,survey4,dtf,idtf,multiskip,sports1m,Simonyan14b,xu2015discriminative,ji,3dcnn}.
Besides detecting action in manually trimmed short video, researchers start to develop techniques for detecting actions in untrimmed long videos in the wild. This trend motivates another challenging topic - \textit{temporal action localization}: given a long untrimmed video, ``when does a specific action start and end?'' This problem is important because real applications usually involve long untrimmed videos, which can be highly unconstrained in space and time, and one video can contain multiple action instances plus background scenes or other activities. Localizing actions in long videos, such as those in surveillance, can save tremendous time and computational costs.

Most state-of-the-art methods rely on manually selected features, and their performances still require much improvement. For example, top performing approaches in THUMOS Challenge 2014 \cite{th1,th2,th3,THUMOS14} and 2015 \cite{th15,THUMOS15} both used improved Dense Trajectory (iDT) with Fisher Vector (FV) \cite{idtf,Oneata2}. There have been some recent attempts at incorporating iDT features with appearance features automatically extracted by frame-level deep networks \cite{th1,th2,th3}. Nevertheless, such 2D ConvNets do not capture motion information, which is important for modeling actions and determining their temporal boundaries.

As an analogy in still images, object detection recently achieved large improvements by using deep networks. Inspired by Region-based Convolutional Neural Networks (R-CNN) \cite{rcnn} and its upgraded versions \cite{fastrcnn,fasterrcnn,deepbox}, we develop Segment-CNN\protect\footnotemark \footnotetext{Source code and trained models are available online at \url{https://github.com/zhengshou/scnn/}.}, which is an effective deep network framework for temporal action localization as outlined in Figure \ref{framework}.
We adopt 3D ConvNets \cite{ji,3dcnn}, which recently has been shown to be promising for capturing motion characteristics in videos, and add a new multi-stage framework.
First, multi-scale segments are generated as candidates for three deep networks. The \textbf{proposal} network classifies each segment as either action or background in order to eliminate background segment estimated to be unlikely to contain actions of interest. The \textbf{classification} network trains typical one-vs-all classification model for all action categories plus the background.

However, the classification network aims at finding key evidences to distinguish different categories, rather than localizing precise action presences in time. Sometimes, the scores from the classification network can be high even when the segment has only a very small overlap with the ground truth instance. This can be detrimental because subsequent post-processing steps, such as Non-Maximum Suppression (NMS), might remove segment of small score but large overlap with ground truth. To explicitly take temporal overlap into consideration, we introduce the \textbf{localization} network based on the same architecture, but this network uses a novel loss function, which rewards segments with higher temporal overlap with the ground truths, and thus can generate confidence scores more suitable for post-processing. Note that the classification network cannot be replaced by the localization network. We will show later that using the trained classification network (without considering temporal overlap) to initialize the localization network (take into account temporal overlap) is important, and achieves better temporal localization accuracies. 


To summarize, our main contributions are three-fold:

(1) To the best of our knowledge, our work is the first to exploit 3D ConvNets with multi-stage processes for temporal action localization in untrimmed long videos in the wild.

(2) We introduce an effective multi-stage Segment-CNN framework, to propose candidate segments, recognize actions, and localize temporal boundaries. The proposal network improves the efficiency by eliminating unlikely candidate segments, and the localization network is key to temporal localization accuracy boosting.

(3) The proposed techniques significantly outperform the state-of-the-art systems over two large-scale benchmarks suitable for temporal action localization. When the overlap threshold used in evaluation is set to 0.5, our approach improves mAP on MEXaction2 from 1.7$\%$ to 7.4$\%$ and mAP  on THUMOS 2014 from 15.0$\%$ to 19.0$\%$. We did not evaluate on THUMOS Challenge 2015 \cite{THUMOS15} because the ground truth is withheld by organizers for future evaluation. More detailed evaluation results are available in Section \ref{exp}.

\section{Related work}

\noindent\textbf{Temporal action localization.} This topic has been studied in two directions. When training data only have video-level category labels but no temporal annotations, researchers formulated this as weakly supervised problems or multiple instance learning problems to learn the key evidences in untrimmed videos and temporally localize actions by selecting key instances \cite{lai1,lai2}. Sun \textit{et al.} \cite{sssn_mm15} transferred knowledge from web images to address temporal localization in untrimmed web videos.

Another line of work focuses on learning from data when the temporal boundaries have been annotated for action instances in untrimmed videos, such as THUMOS. Most of these works pose this as a classification problem and adopt a temporal sliding window approach, where each window is considered as an action candidate subject to classification \cite{Oneata2}. Surveys about action classification methods can be found in \cite{survey1,survey2,survey3,survey4}. Recently, two directions lead the state-of-the-art: (1) Wang \textit{et al.} \cite{dtf} proposed extracting HOG, HOF, MBH features along dense trajectories, and later on they took camera motion into consideration \cite{idtf}. Further improvement can be achieved by stacking features with multiple time skips \cite{multiskip}. (2) Enlighted by the success of CNNs in recent works \cite{alex,Simonyan15}, Karpathy \textit{et al.} \cite{sports1m} evaluated frame-level CNNs on large-scale video classification tasks. Simonyan and Zisserman \cite{Simonyan14b} designed two-stream CNNs to learn from still image and motion flow respectively. In \cite{xu2015discriminative}, a latent concept descriptor of convolutional feature map was proposed, and great results were achieved on event detection with VLAD encoding. To learn spatio-temporal features together, the architecture of 3D ConvNets was explored in \cite{ji,3dcnn}, achieving competitive results. Oneata \textit{et al.} \cite{Oneata} proposed approximately normalized Fisher Vectors to reduce the high dimensionality of FV. Stoian \textit{et al.} \cite{mex1} introduced a two-level cascade to allow fast search for action instances. Instead of precision, these methods focus on improving the efficiency of conventional methods. To specifically address the temporal precision of action detection, Gaidon \textit{et al.} \cite{actoms2,actoms} modeled the structure of action sequence with atomic action units (actoms). The explicit modeling of action units allows for matching more complete action unit sequences, rather than just partial content. However, this requires mannual annotations for actoms, which can be subjective and burdensome. Our paper presented here aims to solve the same problem of precise temporal localization, but without requiring the difficult task of manual annotation for atomic action units.


\hfill\break\noindent\textbf{Spatio-temporal localization.} There have been active explorations about localizing action in space and time simultaneously. Jain \textit{et al.} \cite{tube} and Soomro \textit{et al.} \cite{walk} built their work on supervoxel. Recently, researchers treat this as a tracking problem \cite{learntrack,actiontubes} by leveraging object detectors \cite{objectaction}, especially human detectors \cite{jiang,humanfocus,actiontubes,gangyu} to detect regions of interest in each frame and then output sequences of bounding boxes. Dense trajectories have also been exploited for extracting the action tubes \cite{tubedt,apt}. Jain \textit{et al.} \cite{15000} added object encodings to help action localization.

However, this problem is different from temporal localization, which is the main topic in this paper: (1) When using object detectors to find spatio-temporal regions of interest, such approaches assume that the actions are performed by human or other pre-defined objects. (2) Spatio-temporal localization requires exhaustive annotations for objects of interest on every frame as training data. This makes it overwhelmingly time-consuming particularly for long untrimmed videos compared with the task of simply labeling the start time and end time of an action depicted in the video, which is sufficient to satisfy many applications.

\hfill\break\noindent\textbf{Object detection.} Inspired by the success of deep learning approaches in object detection, we also review R-CNN and its variations. R-CNN consists of selective search, CNN feature extraction, SVM classification, and bounding box regression \cite{rcnn}. Fast R-CNN reshapes R-CNN into a single-stage using multi-task loss, and also has a RoI pooling layer to share the computation of one image in ConvNets \cite{fastrcnn}. Our work differs from R-CNN in the following aspects: (1) Temporal annotations in training videos can be diverse: some are cleanly trimmed action instances cut out from long videos, such as UCF101 \cite{UCF101}, and some are untrimmed long videos but with temporal boundaries annotated for action instances, such as THUMOS \cite{THUMOS14,THUMOS15}. We provide a paradigm that can handle such diverse annotations. (2) As proven in Faster R-CNN \cite{fasterrcnn} which proposes region proposal network, and DeepBox \cite{deepbox} which detects objectness to re-rank the results of R-CNN, using deep networks for learning objectness is effective and efficient. Therefore, we directly use deep network to classify background and action to obtain candidate segments. (3) We remove the regression stage because learning regression for time shift and duration of video segment does not work well in our experiments, probably because actions can be quite diverse, and therefore do not contain consistent patterns for predicting start/end time. To achieve precise localization, we design the localization network using a new loss function to explicitly consider temporal overlap. This can decrease the score for the segment that has small overlap with the ground truth, and increase the segment of larger overlap. This also benefits post-processing steps, such as NMS, to keep segment with higher temporal localization accuracy.

\section{Detailed descriptions of Segment-CNN}\label{approach}

\subsection{Problem setup}

\noindent\textbf{Problem definition.} We denote a video as $X=\left\{ {{x_t}} \right\}_{t = 1}^T$ where $x_t$ is the $t$-th frame in $X$, and $T$ is the total number of frames in $X$. Each video $X$ is associated with a set of temporal action annotations $ \Psi  = \left\{ {\left( {{\psi _m}, \psi _m^{'} ,{k_m}} \right)} \right\}_{m = 1}^M $, where $M$ is the total number of action instances in $X$, and ${k_m}$, ${\psi _m}$, $ \psi _m^{'} $ are, respectively, action category of the instance $m$ and its starting time and ending time (measured by frame ID). ${k_m} \in \left\{ {1, \ldots ,K} \right\}$, where $K$ is the number of categories. During training, we have a set $ {\mathcal T}$ of trimmed videos and a set $\mathcal{U}$ of untrimmed videos. Each trimmed video $X \in {\mathcal T}$ has ${\psi _m} = 1$, $\psi _m^{'}=T$, and $M=1$.

\hfill\break\noindent\textbf{Multi-scale segment generation.} First, each frame is resized to 171 (width) $\times$ 128 (height) pixels. For untrimmed video $X \in  {\cal U}$, we conduct temporal sliding windows of varied lengths as 16, 32, 64, 128, 256, 512 frames with 75\% overlap. For each window, we construct segment $s$ by uniformly sampling 16 frames. Consequently, for each untrimmed video $X$, we generate a set of candidates $\Phi  = \left\{ {\left( {{s_h},{\phi _h},\phi _h^{'}} \right)} \right\}_{h = 1}^H$ as input for the proposal network, where $H$ is the total number of sliding windows for $X$, and ${\phi _m}$ and $ \phi _m^{'} $ are respectively starting time and ending time of the $h$-th segment ${s_h}$. For trimmed video $X \in  {\cal T}$, we directly sample a segment $s$ of 16 frames from $X$ in uniform.

\hfill\break
\noindent\textbf{Network architecture.}\label{arch} 3D ConvNets conducts 3D convolution/pooling which operates in spatial and temporal dimensions simultaneously, and therefore can capture both appearance and motion for action. Given the competitive performances on video classification tasks, our deep networks use 3D ConvNets as the basic architecture in all stages and follow the network architecture of \cite{3dcnn}. All 3D pooling layers use max pooling and have kernel size of 2$\times$2 in spatial with stride 2, while vary in temporal. All 3D convolutional filters have kernel size 3 and stride 1 in all three dimensions. 
Using the notations $\tt conv($number of filters$)$ for the 3D convolutional layer, $\tt pool($temporal kernel size, temporal stride$)$ for the 3D pooling layer, and $\tt fc($number of filters$)$ for the fully connected layer, the layout of these three types of layers in our architecture is as follows: $\tt conv1a($64$)$ - $\tt pool1($1,1$)$ - $\tt conv2a($128$)$ - $\tt pool2($2,2$)$ - $\tt conv3a($256$)$ - $\tt conv3b($256$)$ - $\tt pool3($2,2$)$ - $\tt conv4a($512$)$ - $\tt conv4b($512$)$ - $\tt pool4($2,2$)$ - $\tt conv5a($512$)$ - $\tt conv5b($512$)$ - $\tt pool5($2,2$)$ - $\tt fc6($4096$)$ - $\tt fc7($4096$)$ - $\tt fc8($$K+1$$)$. Each input for this deep network is a segment $s$ of dimension $171\times128\times16$. C3D is training this network on Sports-1M train split \cite{3dcnn}, and we use C3D as the initialization for our proposal and classification networks.

\subsection{Training procedure}

\noindent\textbf{The proposal network}: We train a CNN network ${\Theta _{{\rm{pro}}}}$ as the background segment filter. Basically, $\tt fc8$ has two nodes that correspondingly represent the background (rarely contains action of interest) and being-action (has significant portion belongs to the actions of interest).

We use the following strategy to construct training data ${{\cal S}_{{\rm{pro}}}} = \left\{ {\left( {{s_n},{k_n}} \right)} \right\}_{n = 1}^N$, where label ${k_n} \in \left\{ {0, 1} \right\}$. For each segment of the trimmed video $X \in {\cal T}$, we set its label as positive. For candidate segments from an untrimmed video $X \in {\cal U}$ with temporal annotation $ \Psi$, we assign a label for each segment by evaluating its Intersection-over-Union (IoU) with each ground truth instance in $ \Psi $ : (1) if the highest IoU is larger than 0.7, we assign a positive label; (2) if the highest IoU is smaller than 0.3, we set it as the background. On the perspective of ground truth, if there is no segment that overlaps with a ground truth instance with IoU larger than 0.7, then we assign a positive label segment $s$ if $s$ has the largest IoU with this ground truth and its IoU is higher than 0.5. At last, we obtain ${{\cal S}_{{\rm{pro}}}} = \left\{ {\left( {{s_n},{k_n}} \right)} \right\}_{n = 1}^{N_{\rm{pro}}}$ which consists of all ${N_{\cal T}} + {N_{ {\cal U}}}$ positive segments and ${N_b} \approx {N_{\cal T}} + {N_{ {\cal U}}}$ randomly sampled background segments, where $N_{\rm{pro}} = {N_{\cal T}} + {N_{ {\cal U}}} + {N_b}$. 

In all experiments, we use a learning rate of 0.0001, with the exception of 0.01 for $\tt fc8$, momentum of 0.9, weight decay factor of 0.0005, and drop the learning rate by a factor of 10 for every 10K iterations. The number of total iterations depends on the scale of dataset and will be clarified in Section \ref{exp}.

Note that, compared with the multi-class classification network, this proposal network is simpler because the output layer only consists of two nodes (action or background).

\hfill\break\noindent\textbf{The classification network}: After substantial background segments are removed by the proposal network, we train a classification model ${\Theta _{{\rm{cls}}}}$ for $K$ action categories as well as background.

Preparing the training data ${{\cal S}_{{\rm{cls}}}}$ follows a similar strategy for the proposal network. Except when assigning label for positive segment, the classification network explicitly indicates action category ${k_m} \in \left\{ {1, \ldots ,K} \right\}$. Moreover, in order to balance the number of training data for each class, we reduce the number of background instances to ${N_b} \approx \frac{{{N_{\cal T}} + {N_{ {\cal U}}}}}{K}$.

As for parameters in SGD, the learning rate is 0.0001, with the exception of 0.01 for $\tt fc8$, momentum is 0.9, weight decay factor is 0.0005, and the learning rate is divided by a factor of 2 for every 10K iterations, because the convergence shall be slower when the number of classes increases.


\begin{figure}[b]
\centering
\includegraphics[width=0.5\textwidth]{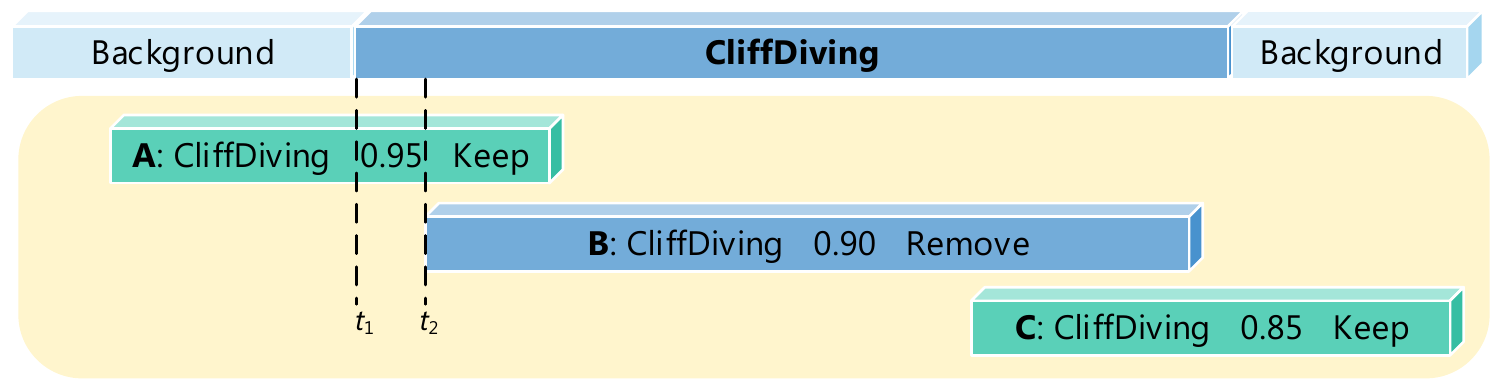}
\caption{Typical case of bad localizations. Assume that the system outputs three predictions: A, B, C. Probably due to that there are some evidences during $[t_1,t_2]$, and A has the highest prediction score. Therefore, the NMS will keep A, remove B, and then keep C. However, actually we hope to remove A and C in NMS, and keep B because B has the largest IoU with the ground truth instance. }
\label{nms}
\end{figure}

\hfill\break\noindent\textbf{The localization network}: As illustrated in Figure \ref{nms}, it is important to push up the prediction score of the segment with larger overlap with the ground truth instance and decrease the scores of the segment with smaller overlap, to make sure that the subsequent post-processing steps can choose segments with higher overlap over those with small overlap. Therefore, we propose this localization network ${\Theta _{{\rm{loc}}}}$ with a new loss function, which takes IoU with ground truth instance into consideration.

Training data ${{\cal S}_{{\rm{loc}}}}$ for the localization network are augmented from ${{\cal S}_{{\rm{cls}}}}$ by associating each segment ${s}$ with the measurement of overlap,  ${v}$. In specific, we set ${v}=1$ for $s$ from trimmed video. If $s$ comes from untrimmed video and has positive label $k$, we set ${v}$ equal to the overlap (measured by IoU) of segment $s$ with the associated ground truth instance. If $s$ is a background segment, as we can see later, its overlap measurement $v$ will not affect our new loss function and gradient computation in back-propagation, and thus we simply set its $v$ as 1.

During each mini-batch, we have $N$ training samples $\left\{ {\left( {{s_n},{k_n},{v_n}} \right)} \right\}_{n = 1}^N$. For the $n$-th segment, the output vector of $\tt fc8$ is $O_n$ and the prediction score vector after the softmax layer is $P_n$. Note that for the $i$-th class, $P_n^{\left( i \right)} = \frac{{{e^{O_n^{\left( i \right)}}}}}{{\sum\nolimits_{j = 1}^N {{e^{O_n^{\left( {j} \right)}}}} }}$. The new loss function is formed by combining ${{\cal L}_{{\rm{softmax}}}}$ and  ${{\cal L}_{{\rm{overlap}}}}$ : \begin{equation}
{{\cal L}} = {{\cal L}_{{\rm{softmax}}}} + \lambda  \cdot {{\cal L}_{{\rm{overlap}}}}
,\end{equation} where $\lambda$ balances the contribution from each part, and through empirical validation, we find that $\lambda=1$ works well in practice. ${{\cal L}_{{\rm{softmax}}}}$ is the conventional softmax loss and is defined as \begin{equation} {{\cal L}_{{\rm{softmax}}}} = \frac{1}{N}\sum\limits_n {\left( { - \log \left( {P_n^{\left( {{k_n}} \right)}} \right)} \right)} , \end{equation} which is effective for training deep networks for classification. ${{\cal L}_{{\rm{overlap}}}}$ is designed to jointly reduce the classification error and adjust the intensity of confidence score according to the extent of overlap: \begin{equation}
{{\cal L}_{{\rm{overlap}}}} = \frac{1}{N}\sum\limits_n {\left( {\frac{1}{2} \cdot \left( {\frac{{{{\left( {P_n^{\left( {{k_n}} \right)}} \right)}^2}}}{{{{\left( {{v_n}} \right)}^\alpha }}} - 1} \right) \cdot \left[ {{k_n} > 0} \right]} \right)} .
\end{equation} Here, $\left[ {{k_n} > 0} \right]$ is equal to 1 when the true class label $k_n$ is positive, and it is equal to 0 when $k_n=0$, which means the $s_n$ is a background training sample. ${{\cal L}_{{\rm{overlap}}}}$ is intended to boost the detection scores ($P$) of segments that have high overlaps ($v$) with ground truth instances, and reduce the scores of those with small overlaps. The hyper-parameter $\alpha$ controls the adjustment range for the intensity of the confidence score. The sensitivity of $\alpha$ is explored in Section \ref{exp}. In addition, the total gradient w.r.t output of the $i$-th node in $\tt fc8$ is as follows: \begin{equation}
\frac{{\partial {{\cal L}}}}{{\partial O_n^{\left( i \right)}}} = \frac{{\partial {{\cal L}_{{\rm{softmax}}}}}}{{\partial O_n^{\left( i \right)}}} + \lambda  \cdot \frac{{\partial {{\cal L}_{{\rm{overlap}}}}}}{{\partial O_n^{\left( i \right)}}},
\end{equation} in which \begin{equation}
\frac{{\partial {{\cal L}_{{\rm{softmax}}}}}}{{\partial O_n^{\left( i \right)}}} = \left\{ {\begin{array}{*{20}{c}}
{\frac{1}{N} \cdot \left( {P_n^{\left( {{k_n}} \right)} - 1} \right)}&{{\rm{if \: }}i = {k_n}}\\
{\frac{1}{N} \cdot P_n^{\left( i \right)}}&{{\rm{if\: }} i \ne {k_n}}
\end{array}} \right.
\end{equation} and \begin{equation}
\frac{{\partial {{\cal L}_{{\rm{overlap}}}}}}{{\partial O_n^{\left( i \right)}}} = \left\{ {\begin{array}{*{20}{r}}
\multicolumn{1}{c}{\frac{1}{N} \cdot \left( {\frac{{{{\left( {P_n^{\left( {{k_n}} \right)}} \right)}^2}}}{{{{\left( {{v_n}} \right)}^\alpha }}} \cdot \left( {1 - P_n^{\left( {{k_n}} \right)}} \right)} \right) \cdot \left[ {{k_n} > 0} \right]}\\
{{\rm{if\: }}i = {k_n}}\\
\multicolumn{1}{c}{\frac{1}{N} \cdot \left( {\frac{{{{\left( {P_n^{\left( {{k_n}} \right)}} \right)}^2}}}{{{{\left( {{v_n}} \right)}^\alpha }}} \cdot \left( { - P_n^{\left( i \right)}} \right)} \right) \cdot \left[ {{k_n} > 0} \right]}\\
{{\rm{if\: }}i \ne {k_n}}
\end{array}} \right. . \end{equation}

\begin{figure}[t]
\centering
\includegraphics[width=0.42\textwidth]{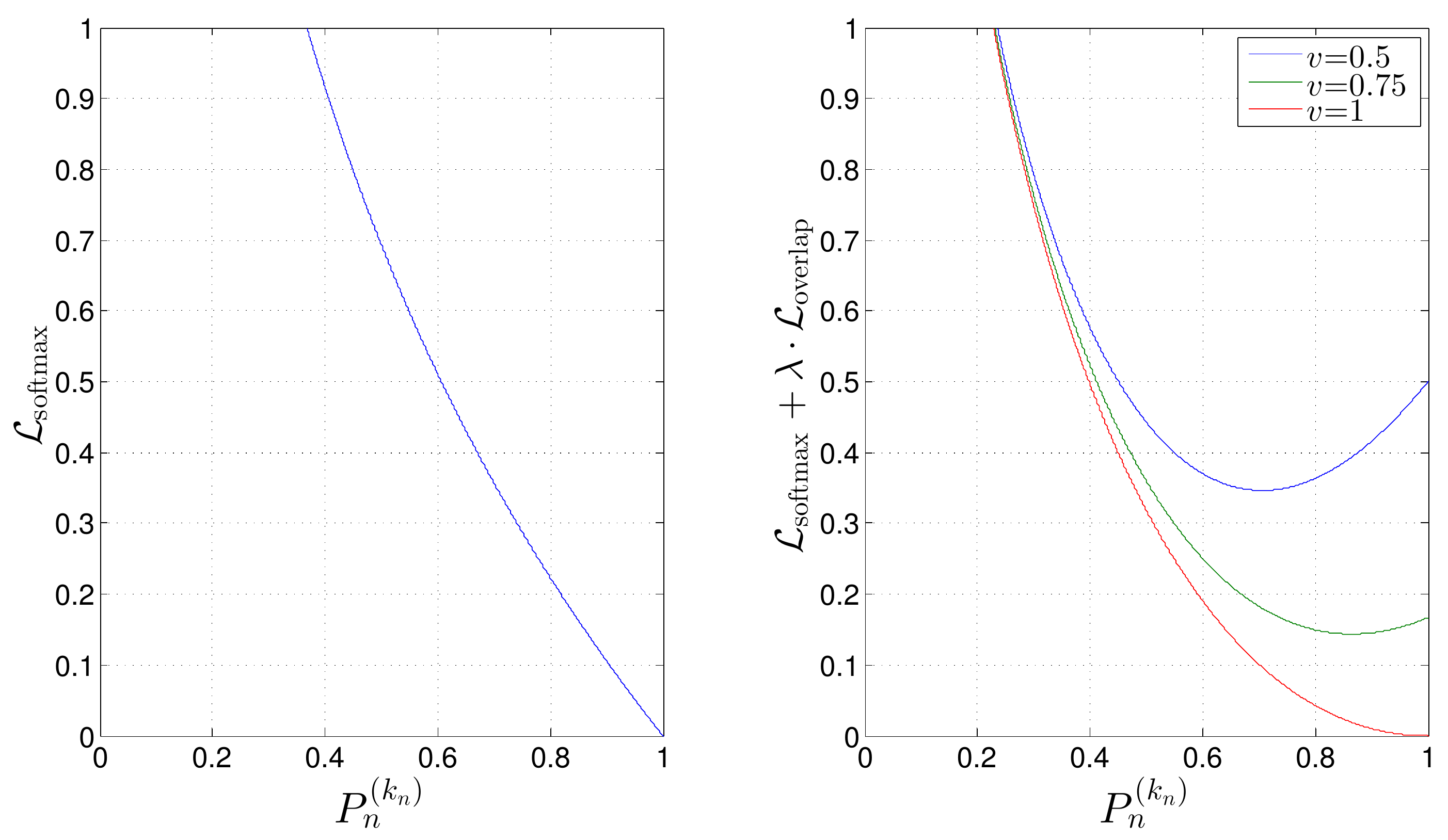}
\caption{An illustration of how ${{\cal L}_{{\rm{overlap}}}}$ works compared with ${{\cal L}_{{\rm{softmax}}}}$ for each positive segment. Here we use $\alpha=1$, $\lambda=1$, and vary overlap $v$ in ${{\cal L}_{{\rm{overlap}}}}$. The $x$-axis is the prediction score at the node that corresponds to true label, and the $y$-axis is the loss.}
\label{loss}
\end{figure}

Given a training sample ${\left( {{s_n},{k_n},{v_n}} \right)}$, Figure \ref{loss} shows how ${{\cal L}_{{\rm{overlap}}}}$ influences the original softmax loss. It also provides more concrete insights about the design of this loss function. (1) If the segment belongs to the background, ${{\cal L}_{{\rm{overlap}}}}=0$ and ${{\cal L}}={{\cal L}_{{\rm{softmax}}}}$. (2) If the segment is positive, ${\cal L}$ reachs the minimum at $P_n^{\left( {{k_n}} \right)} = \sqrt {{{\left( {{v_n}} \right)}^\alpha }} $, and therefore penalizes two cases: either $P_n^{\left( {{k_n}} \right)}$ is too small due to misclassification, or $P_n^{\left( {{k_n}} \right)}$ explodes and exceeds the learning target $\sqrt {{{\left( {{v_n}} \right)}^\alpha }}$ which is proportional to overlap $v_n$. Also note that ${{\cal L}}$ is designed to increase as $v_n$ decreases, considering that the training segment with smaller overlap with ground truth instance is less reliable because it may include considerable noise. (3) In particular, if this positive segment has overlap $v_n = 1$, the loss function becomes similar to the softmax loss, and ${{\cal L}}$ gradually decreases from $ + \infty $ to 1 as ${P_n^{\left( {{k_n}} \right)}}$ goes from 0 to 1.

During optimization, ${\Theta _{{\rm{loc}}}}$ is fine-tuned on ${\Theta _{{\rm{cls}}}}$. Because doing classification is also one objective of the localization network, and a trained classification network can be good initialization. We use the same learning rate, momentum, and weight decay factor as for the classification network. Other parameters depending on the dataset are indicated in Section \ref{exp}.


\subsection{Prediction and post-processing}

During prediction, we slide varied length temporal window to generate a set of segments and input them into ${\Theta _{{\rm{pro}}}}$ to obtain proposal scores ${ {P}_{{\rm{pro}}}}$. In this paper, we keep segments with ${ {P}_{{\rm{pro}}}} \ge 0.7$. Then we evaluate the retained segments by ${\Theta _{{\rm{loc}}}}$ to obtain action category predictions and confidence scores ${ {P}_{{\rm{loc}}}}$. During post-processing, we remove all segments predicted as the background and refine ${ {P}_{{\rm{loc}}}}$ by multiplying with class-specific frequency of occurrence for each window length in the training data to leverage window length distribution patterns. Finally, because redundant detections are not allowed in evaluation, we conduct NMS based on ${ {P}_{{\rm{loc}}}}$ to remove redundant detections, and set the overlap threshold in NMS to a little bit smaller than the overlap threshold $\theta$ in evaluation ($\theta - 0.1$ in this paper).

\section{Experiments} \label{exp}

\subsection{Datasets and setup}

\noindent\textbf{MEXaction2 \cite{mex2}.} This dataset contains two action classes: ``BullChargeCape'' and ``HorseRiding''. This dataset consists of three subsets: INA videos, YouTube clips, and UCF101 Horse Riding clips. YouTube clips and UCF101 Horse Riding clips are trimmed, whereas INA videos are untrimmed and are approximately 77 hours in total. With regard to action instances with temporal annotation, they are divided into train set (1336 instances), validation set (310 instances), and test set (329 instances).

\begin{figure*}[t]
\centering
\includegraphics[width=\textwidth]{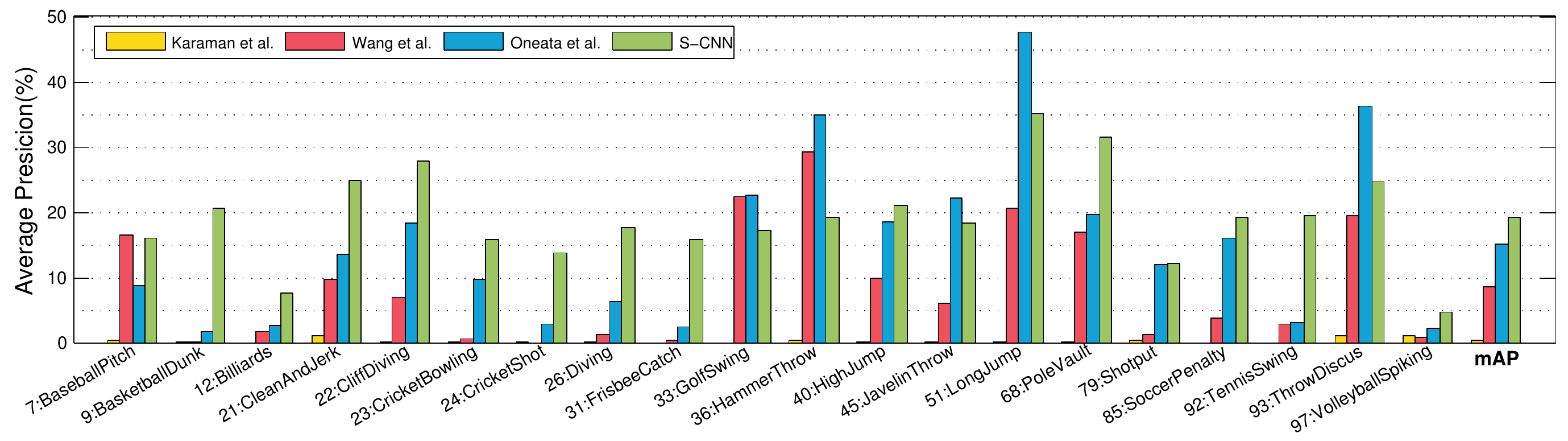}
\caption{Histogram of average precision (\%) for each class on THUMOS 2014 when the overlap threshold is set to 0.5 during evaluation.}
\label{map3}
\end{figure*}

\hfill\break\noindent\textbf{THUMOS 2014 \cite{THUMOS14}.} The temporal action detection task in THUMOS Challenge 2014 is dedicated to localizing action instances in long untrimmed videos. The detection task involves 20 categories as indicated in Figure \ref{map3}. The trimmed videos used for training are 2755 videos of these 20 actions in UCF101. The validation set contains 1010 untrimmed videos with temporal annotations of 3007 instances in total. The test set contains 3358 action instances from 1574 untrimmed videos, whereas only 213 of them contain action instances of interest. We exclude the remaining 1361 background videos in the test set.

\subsection{Comparison with state-of-the-art systems}\label{compare}

\noindent\textbf{Evaluation metrics.} \label{eval} We follow the conventional metrics used in THUMOS Challenge to regard temporal action localization as a retrieval problem, and evaluate average precision (AP). A prediction is marked as correct only when it has the correct category prediction, and has IoU with ground truth instance larger than the overlap threshold (measured by IoU). Note that redundant detections are not allowed.

\hfill\break\noindent\textbf{Results on MEXaction2.} We build our system based on Caffe \cite{caffe} and C3D \cite{3dcnn}. We use the train set in MEXaction2 for training. The number of training iterations is 30K for the proposal network, 20K for the classification network, and 20K in the localization network with $\alpha=0.25$.

\begin{figure*}[t]
\centering
\includegraphics[width=\textwidth]{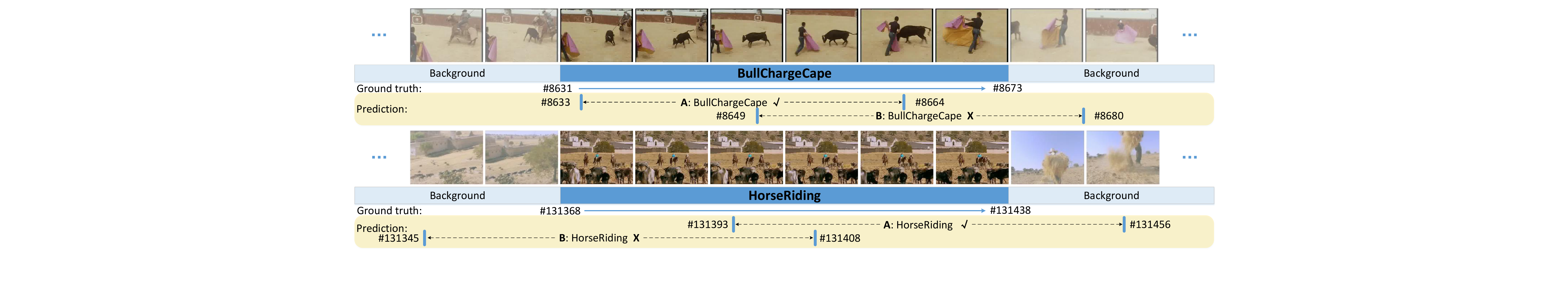}
\caption{Prediction results for two action instances from MEXaction2 when the overlap threshold is set to 0.5 during evaluation. For each ground truth instance, we show two prediction results: A has the highest confidence score among the predictions associated with this ground truth, and B is an incorrect prediction. BullChargeCape: A is correct, but B is incorrect because each ground truth only allows one detection. HorseRiding: A is correct, but B is incorrect because each ground truth only allows one detection. The numbers shown with \# are frame IDs.}
\label{show1}
\end{figure*}

We denote our Segment-CNN using the above settings as S-CNN and compare with typical dense trajectory features (DTF) with bag-of-visual-words representation. The results of DTF is provided by \cite{mex2} \protect\footnotemark \footnotetext{Note that the results reported in \cite{mex2} use different evaluation metrics. To make them comparable, we re-evaluate their prediction results according to standard criteria mentioned in Section \ref{eval}. }, which trains three SVM models with different set of negative samples and averages AP overall.
According to Table \ref{map1}, our Segment-CNN achieves tremendous performance gain for ``BullChargeCape'' action and competitive performance for  ``HorseRiding" action. Figure \ref{show1} displays our prediction results for ``BullChargeCape'' and ``HorseRiding'', respectively.

\begin{table}[h]
\begin{center}
\begin{tabular}{c|cc|c}
AP$(\%)$ & BullChargeCape & HorseRiding & mAP \\ \hline
DTF &       0.3        &        3.1     &   1.7  \\ \hline
S-CNN   &       11.6         &    3.1         &   7.4
\end{tabular}
\end{center}
\caption{Average precision on MEXaction2. The overlap threshold is set to 0.5 during evaluation.}
\label{map1}
\end{table}

\hfill\break\noindent\textbf{Results on THUMOS 2014\protect\footnotemark}: \footnotetext{Note that the evaluation toolkit used in THUMOS 2014 has some bugs, and recently the organizers released a new toolkit with fair evaluation criteria. Here, we re-evaluate the submission results of all teams using the updated toolkit.} The instances in train set and validation set are used for training. The number of training iterations is 30K for all three networks. We again set $\alpha=0.25$ for the localization network. We denote our Segment-CNN using the above settings as S-CNN.

\begin{table}[h]
\begin{center}
\begin{tabular}{c|ccccc}
$\theta$ & 0.1 & 0.2 & 0.3 & 0.4 & 0.5 \\ \hline
Karaman et al. \cite{th3} &  1.5 &  0.9 &  0.5 &  0.3 &  0.2   \\
Wang et al. \cite{th2} &  19.2 &  17.8 &  14.6 &  12.1 &  8.5   \\
Oneata et al. \cite{th1} &  39.8 &  36.2 &  28.8 &  21.8 &  15.0   \\ \hline
S-CNN  &  47.7 &  43.5 &  36.3 &  28.7 & 19.0
\end{tabular}
\end{center}
\caption{Mean average precision on THUMOS 2014 as the overlap IoU threshold $\theta$ used in evaluation varies.}
\label{map2}
\end{table}

\begin{figure*}[t]
\centering
\includegraphics[width=\textwidth]{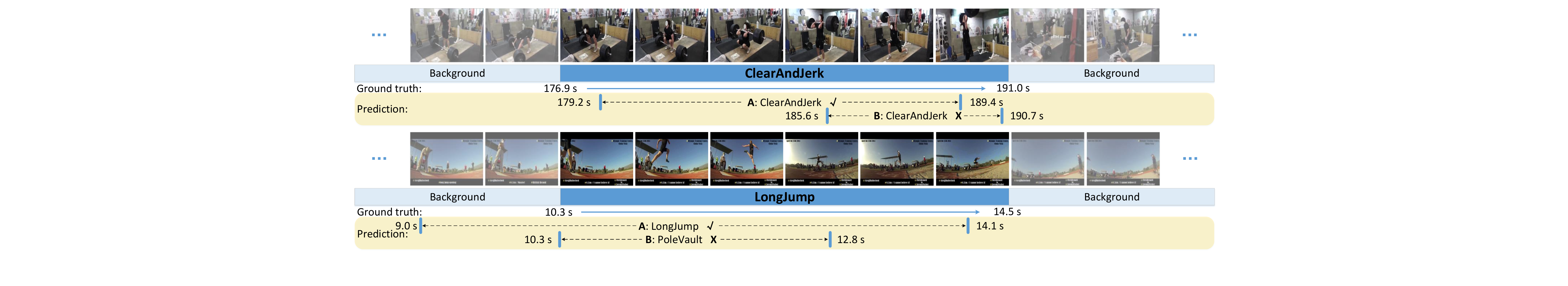}
\caption{Prediction results for two action instances from THUMOS 2014 test set when the overlap threshold is set to 0.5 during evaluation. For each ground truth instance, we show two prediction results: A has the highest confidence score among the predictions associated with this ground truth, and B is an incorrect prediction. ClearAndJerk: A is correct, but B is incorrect because its overlap IoU with ground truth is less than threshold 0.5. LongJump: A is correct, but B is incorrect because it has the wrong action category prediction - PoleVault.}
\label{show2}
\end{figure*}

As for comparisons, beyond DTF, several baseline systems incorporate frame-level deep networks and even utilize lots of other features: (1) Karaman et al. \cite{th3} used FV encoding of iDT with weighted saliency based pooling, and conducted late fusion with frame-level CNN features. (2) Wang et al. \cite{th2} built a system on iDT with FV representation and frame-level CNN features, and performed post-processing to refine the detection results. (3) Oneata et al. \cite{th1} conducted localization using FV encoding of iDT on temporal sliding windows, and performed post-processing following \cite{Oneata2}. Finally, they conducted weighted fusion for the localization scores of temporal windows and video-level scores generated by classifiers trained on iDT features, image features, and audio features.
The results are listed in Table \ref{map2}. AP for each class can be found in Figure \ref{map3}. Our Segment-CNN significantly outperforms other systems for 14 of 20 actions, and the average performance improves from 15.0\% to 19.0\%. We also show two prediction results for the THUMOS 2014 test set in Figure \ref{show2}.

\hfill\break\noindent\textbf{Efficiency analysis.} Our approach is very efficient when compared with all other systems, which typically fuse different features, and therefore can become quite cumbersome. Most segments generated from sliding windows are removed by the first proposal network, and thus the operations in classification and localization are greatly reduced. For each batch, the speed is around 1 second, and the number of segments can be processed during each batch depends on the GPU memory (approximately 25 for GeForce GTX 980 of 4G memory). The storage requirement is also extremely small because our method does not need to cache intermediate high dimensional features, such as FV to train SVM. All required by Segment-CNN is three deep network models, which occupy less than 1 GB in total.

\subsection{Impact of individual networks}

To study the effects of each network individually, we compare four Segment-CNNs using different settings: (1) \textbf{S-CNN}: keep all three networks and settings in Section \ref{compare}, and ${\Theta _{{\rm{loc}}}}$ is fine-tuned on ${\Theta _{{\rm{cls}}}}$; (2) \textbf{S-CNN (w/o proposal)}: remove the proposal network completely, and directly use ${\Theta _{{\rm{loc}}}}$ to do predictions on sliding windows; (3) \textbf{S-CNN (w/o classification)}: remove the classification network completely and thus do not have ${\Theta _{{\rm{cls}}}}$ to serve as initialization for training ${\Theta _{{\rm{loc}}}}$; (4) \textbf{S-CNN (w/o localization)}: remove the localization network completely and instead use classification model ${\Theta _{{\rm{cls}}}}$ to produce predictions.

\hfill\break\noindent\textbf{The proposal network.} We compare S-CNN (w/o proposal) and S-CNN, which includes the proposal network as described above (two nodes in $\tt fc8$). Because of the smaller network architecture than S-CNN (w/o proposal), S-CNN can reduce the number of operations conducted on background segments, and therefore accelerate speed. In addition, the results listed in Table \ref{scale} demonstrate that keeping the proposal network can also improve precision because it is designed for filtering out background segments that lack action of interests.

\begin{table}[h]
\begin{center}
\begin{tabular}{c|c|c}
networks     & S-CNN (w/o proposal) & S-CNN  \\ \hline
mAP$(\%)$         &  17.1     &     19.0        \\ 
\end{tabular}
\end{center}
\caption{mAP comparisons on THUMOS 2014 between removing the proposal network and keeping the proposal network. The overlap threshold is set to 0.5 during evaluation.}
\label{scale}
\end{table}

\begin{figure}[h]
\centering
\includegraphics[width=0.45\textwidth]{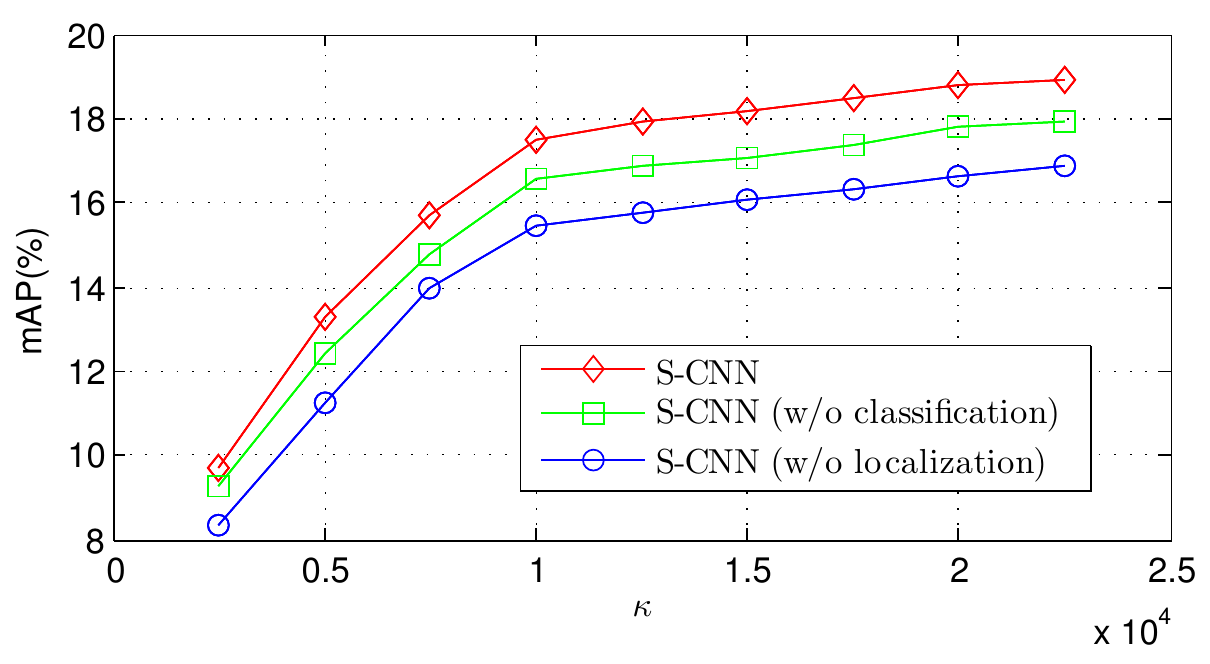}
\caption{Effects of the classification and localization networks. $y$-axis is mAP(\%) on THUMOS 2014, and $x$-axis varies the depth $\kappa$ in top-$\kappa$ selection. The overlap threshold is set to 0.5 during evaluation.}
\label{top-k}
\end{figure}

\noindent\textbf{The classification network.} Although ${\Theta _{{\rm{cls}}}}$ is not used during prediction, the classification network is still important because fine-tuning on ${\Theta _{{\rm{cls}}}}$ results in better performance. During evaluation here, we perform top-$\kappa$ selection on the final prediction results to select $\kappa$ segments with maximum confidence scores. As shown in Figure \ref{top-k}, S-CNN fine-tuned on ${\Theta _{{\rm{cls}}}}$ outperforms S-CNN (w/o classification) consistently when $\kappa$ varies, and consequently the classification network is necessary during training.

\hfill\break\noindent\textbf{The localization network.}\label{localization} Figure \ref{top-k} also proves the effectiveness of the localization network. By adding the localization network, S-CNN can significantly improve performances compared with the baseline S-CNN (w/o localization), which only contains the proposal and classification networks. This is because the new loss function introduced in the localization network refines the scores in favoring segments of higher overlap with the ground truths, and therefore higher temporal localization accuracy can be achieved.

In addition, we vary $\alpha$ in the overlap loss term ${{\cal L}_{{\rm{overlap}}}}$ of the loss function to evaluate its sensitivity. We find that our approach has stable performances over a range of $\alpha$ value (e.g., from 0.25 to 1.0).

\section{Conclusion}

We propose an effective multi-stage framework called Segment-CNN to address temporal action localization in untrimmed long videos. Through the above evaluation for each network, we demonstrate the contribution from the proposal network to identify candidate segments, the necessity of the classification network to provide good initialization for training the localization model, and the effectiveness of the new loss function used in the localization network to precisely localize action instances in time.

In the future, we would like to extend our work to events and activities, which usually consist of multiple actions, therefore precisely localizing action instances in time can be helpful for their recognition and detection.

\section{Acknowledgment}


This work is supported by the Intelligence Advanced Research Projects Activity (IARPA) via Department of Interior National Business Center contract number D11PC20071. The U.S. Government is authorized to reproduce and distribute reprints for Governmental purposes notwithstanding any copyright annotation thereon. Disclaimer: The views and conclusions contained herein are those of the authors and should not be interpreted as necessarily representing the official policies or endorsements, either expressed or implied, of IARPA, DOI-NBC, or the U.S. Government. We thank Dong Liu, Guangnan Ye, and anonymous reviewers for the insightful suggestions.

{\small
\bibliographystyle{ieee}
\bibliography{egbib}
}

\end{document}